\documentclass[11pt,a4paper]{article}

\usepackage{amssymb}
\usepackage{amsmath}
\usepackage{amstext}
\usepackage{graphicx}
\usepackage{subfigure}
\usepackage{url}

\usepackage[T1]{fontenc}
\usepackage[utf8]{inputenc}
\usepackage{authblk}
\usepackage[resetlabels]{multibib}

\title{Accurate Pulmonary Nodule Detection in Computed Tomography Images Using Deep Convolutional Neural Networks}

\author[]{Jia Ding$^*$, Aoxue Li$^*$, Zhiqiang Hu \thanks{dingjia@pku.edu.cn, lax@pku.edu.cn, huzq@pku.edu.cn; \\\qquad\qquad These authors have contributed equally to this work.}}
\author[]{Liwei Wang\thanks{\textbf{wanglw@cis.pku.edu.cn, corresponding author}}}
\affil[]{School of EECS, Peking University}

\date{}

\begin{document}

\maketitle

\begin{abstract}
Early detection of pulmonary cancer is the most promising way to
enhance a patient's chance for survival. Accurate pulmonary nodule
detection in computed tomography (CT) images is a crucial step in
diagnosing pulmonary cancer. In this paper, inspired by the
successful use of deep convolutional neural networks (DCNNs) in
natural image recognition, we propose a novel pulmonary nodule
detection approach based on DCNNs. We first introduce a
deconvolutional structure to Faster Region-based Convolutional
Neural Network (Faster R-CNN) for candidate detection on axial
slices. Then, a three-dimensional DCNN is presented for the
subsequent false positive reduction. Experimental results of the
LUng Nodule Analysis 2016 (LUNA16) Challenge demonstrate the
superior detection performance of the proposed approach on nodule
detection(average FROC-score of 0.891, ranking the \textbf{1st
place} over all submitted results).
\end{abstract}
\section{Introduction}
Pulmonary cancer, causing 1.3 million deaths annually, is a leading
cause of cancer death worldwide \cite{cancer}. Detection and
treatment at an early stage are required to effectively overcome
this burden. Computed tomography (CT) was recently adopted as a
mass-screening tool for pulmonary cancer diagnosis, enabling rapid
improvement in the ability to detect tumors early. Due to the
development of CT scanning technologies and rapidly increasing
demand, radiologists are overwhelmed with the amount of data they
are required to analyze.

Computer-Aided Detection (CAD) systems have been developed to assist
radiologists in the reading process and thereby potentially making
pulmonary cancer screening more effective. The architecture of a CAD
system for pulmonary nodule detection typically consists of two
stages: nodule candidate detection and false positive reduction.
Many CAD systems have been proposed for nodule detection
\cite{Torres15,luna}. Torres et al. detect candidates with a
dedicated dot-enhancement filter and then a feed-forward neural
network based on a small set of hand-craft features is used to
reduce false positives \cite{Torres15}.
Although conventional CAD systems have yielded promising results,
they still have two distinct drawbacks as follows.
\begin{itemize}
\item Traditional CAD systems detect candidates based on some simple assumptions
(eg. nodules look like a sphere) and propose some low-level
descriptors \cite{Torres15}. Due to the high variability of
nodule shape, size, and texture, low-level descriptors fail to
capture discriminative features, resulting in inferior detection
results.
\item Since CT images are 3D inherently, 3D contexts play an important role in recognizing nodules.
However, several 2D/2.5D deep neural networks have achieved
promising performance in false positive reduction
\cite{Setio16,Zagoruyko16} while rare works focus on introducing 3D
contexts for nodule detection directly.
\end{itemize}

\begin{figure}[t]
\vspace{0.02in}
\begin{center}
\includegraphics[width=\columnwidth]{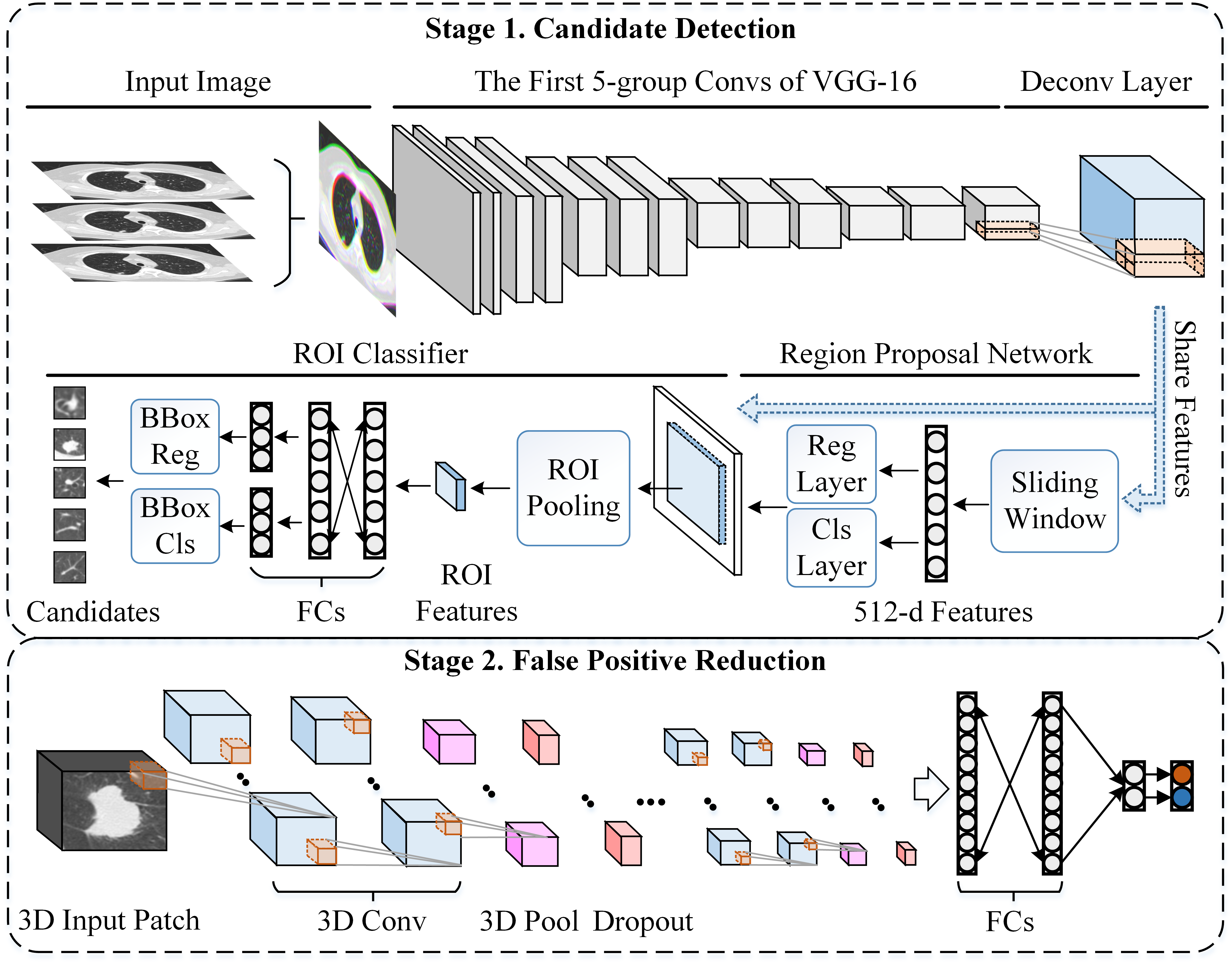}
\end{center}
\vspace{-0.2in} \caption{The framework of the proposed CAD system.}
\label{fig:framework}
\end{figure}

In this paper, to address the aforementioned two issues, we propose
a novel CAD system based on DCNNs for accurate pulmonary nodule
detection. In the proposed CAD system, we first introduce a
deconvolutional structure to Faster Region-based Convolutional
Neural Network (Faster R-CNN), the state-of-the-art general object
detection model, for candidate detection on axial slices. Then, a
three-dimensional DCNN (3D DCNN) is presented for false positive
reduction. The framework of our CAD system is illustrated in Fig.
\ref{fig:framework}.

To evaluate the effectiveness of our CAD system, we test it on the
LUng Nodule Analysis 2016 (LUNA16) challenge \cite{luna}, and yield
the 1st place of Nodule Detection Track (NDET) with an average
FROC-score of 0.891. Our system achieves high
detection sensitivities of 92.2$\%$ and 94.4$\%$ at 1 and 4 false
positives per scan, respectively.
\section{The proposed CAD system}
In this section, we propose a CAD system based on DCNNs for accurate
pulmonary nodule detection, where two main stages are incorporated:
(1) candidate detection by introducing the deconvolutional structure
into Faster R-CNN and (2) false positive reduction by using a
three-dimensional DCNN.
\subsection{Candidate Detection Using Improved Faster R-CNN}
\label{sect:FasterRCNN}

Candidate detection, as a crucial step in the CAD systems, aims to
restrict the total number of nodule candidates while remaining high
sensitivity. Inspired by the successful use of DCNNs in object
recognition \cite{faster_rcnn}, we propose a DCNN model for
detecting nodule candidates from CT images, where the
deconvolutional structure is introduced into the state-of-the-art
general object detection model, Faster R-CNN, for fitting the size
of nodules. The details of our candidate detection model are given
as follows.

We first describe the details of generating inputs for our candidate
detection network. Since using 3D volume of original CT scan as the
DCNN input gives rise to high computation cost, we use axial slices
as inputs instead. For each axial slice in CT images, we concatenate
its two neighboring slices in axial direction, and then rescale it
into $600\times600\times3$ pixels (as shown in Fig.
\ref{fig:framework}).

In the following, we describe the details of the architecture of the
proposed candidate detection network. The network is composed of two
modules: a region proposal network (RPN) aims to propose potential
regions of nodules (also called Region-of-Interest (ROI)); a ROI
classifier then recognizes whether ROIs are nodules or not. In order
to save computation cost of training DCNNs, these two DCNNs share
the same feature extraction layers.

\noindent\textbf{Region Proposal Network} The region proposal
network takes an image as input and outputs a set of rectangular
object proposals (i.e. ROIs), each with an objectness score
\cite{faster_rcnn}. The structure of the network is given as
follows.

Owing to the much smaller size of pulmonary nodules compared with
common objects in natural images, original Faster R-CNN, which
utilizes five-group convolutional layers of VGG-16Net \cite{VGG} for
feature extraction, cannot explicitly depict the features of nodules
and results in a limited performance in detecting ROIs of nodules.
To address this problem, we add a deconvolutional layer, whose
kernel size, stride size, padding size and kernel number are 4, 4, 2 and 512
respectively, after the last layer of the original feature
extractor. Note that, the added deconvolutional layer recovers more
fine-grained features compared with original feature maps, our
model thus yields much better detection results than the
original Faster R-CNN. To generate ROIs, we slide a small network
over the feature map of the deconvolutional layer. This small
network takes a $3\times3$ spatial window of deconvolutional feature
map as input and map each sliding window to a 512-dimensional
feature. The feature is finally fed into two sibling fully-connected
layers for regressing the boundingbox of regions (i.e. Reg Layer in
Fig. \ref{fig:framework}) and predicting objectness score (i.e. Cls
Layer in Fig. \ref{fig:framework}), respectively.

\begin{figure}[htbp]
\vspace{-0.2in}
\begin{center}
\includegraphics[width=0.5\columnwidth]{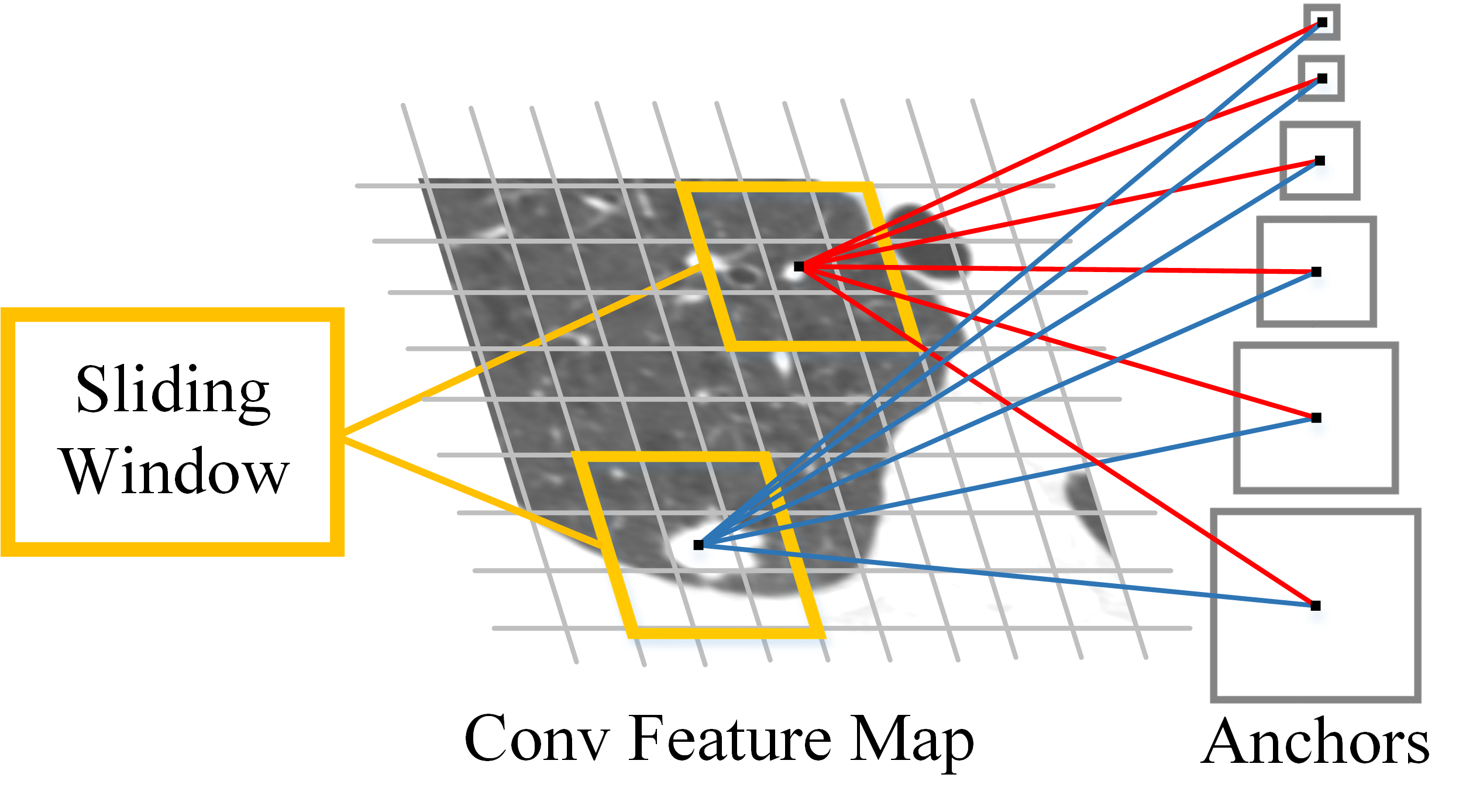}
\end{center}
\vspace{-0.2in} \caption{Illustration of anchors in the improved
Faster R-CNN} \vspace{-0.2in}\label{fig:anchor}
\end{figure}

At each sliding-window location, we simultaneously predict multiple
ROIs. The multiple ROIs are parameterized relative to the
corresponding reference boxes, which we call anchors. To fit the
size of nodules, we design six anchors with different size for each
sliding window: $4\times4$, $6\times6$, $10\times10$, $16\times16$,
$22\times22$, and $32\times32$ (See Fig. \ref{fig:anchor}). The
detailed description of RPN is given in \cite{faster_rcnn}.

\noindent\textbf{ROI Classification Using Deep Convolutional Neural
Network} With the ROIs extracted by RPN, a DCNN is developed to
decide whether each ROI is nodule or not. A ROI Pooling layer is
firstly exploited to map each ROI to a small feature map with a
fixed spatial extent $W\times H$ ($7\times7$ in this paper). The ROI
pooling works by dividing the ROI into an $W\times H$ grid of
sub-windows and then max-pooling the values in each sub-window into
the corresponding output grid cell. Pooling is applied independently
to each feature map channel as in standard max pooling. After ROI
pooling layer, a fully-connected network, which is composed of two
4096-way fully-connected layers, then map the fixed-size feature map
into a feature vector. A regressor and a classifier based on the
feature vector (i.e. BBox Reg and BBox Cls in
Fig.\ref{fig:framework}) then respectively regress boundingboxes of
candidates and predict candidate confidence scores.

In the training process, by merging the RPN and ROI classifier into
one network, we define the loss function for an image as follows.
\begin{equation}
\mathcal{L}_t= \frac{1}{N_{c}}\sum_{i}\mathcal
{L}_{c}(\hat{p}_i,p^*_i)+\frac{1}{N_{r}}\sum_{i}\mathcal
{L}_{r}(\hat{t}_i,t^*_i)+\frac{1}{N_{c'}}\sum_{j}\mathcal
{L}_{c}(\tilde{p}_j,p^*_j)+\frac{1}{N_{r'}}\sum_{j}\mathcal
{L}_{r}(\tilde{t}_j,t^*_j)
\end{equation}
where $N_{c}$, $N_{r}$, $N_{c'}$ and $N_{r'}$ denote the total
number of inputs in Cls Layer, Reg Layer, BBox Cls and BBox Reg,
respectively. The $\hat{p}_i$ and $p^*_i$ respectively denote the
predicted and true probability of anchor $i$ being a nodule.
$\hat{t}_i$ is a vector representing the 4 parameterized coordinates
of the predicted bounding box of RPN, and $t^*_i$ is that of the
ground-truth box associated with a positive anchor. In the same way,
$\tilde{p}_j$, $p^*_j$, $\hat{t}_j$ and $t^*_j$ denote the
corresponding concepts in the ROI classifier. The detailed
definitions of classification loss $\mathcal {L}_c$ and regression
loss $\mathcal {L}_r$ are the same as the corresponding definitions
in the literature \cite{faster_rcnn}.
\subsection{False Positive Reduction Using 3D DCNN}
\begin{figure}[htbp]
\vspace{0.02in}
\begin{center}
\includegraphics[width=\columnwidth]{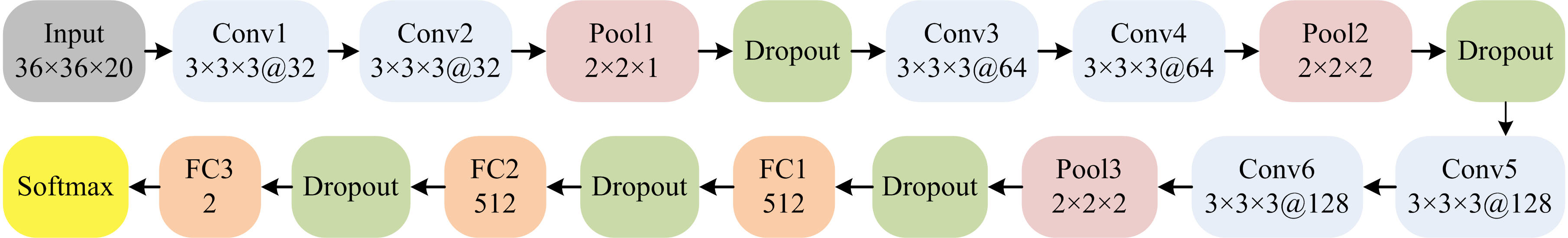}
\end{center}
\vspace{-0.15in} \caption{The architecture of the proposed
three-dimensional deep convolutional neural network. In this figure,
`Conv', `Pool', and `FC' denote the convolutional layer, pooling
layer and fully-connected layer, respectively.} \label{fig:3dcnn}
\end{figure}

In the consideration of time and space cost, we propose a
two-dimensional (2D) DCNN (i.e. Improved Faster R-CNN) to detect
nodule candidates (See Section \ref{sect:FasterRCNN}). With the
extracted nodule candidates, a 3D DCNN, which captures the full
range of contexts of candidates and generates more discriminative
features compared with 2D DCNNs, is utilized for false positive
reduction. This network contains six 3D convolutional layers which
are followed by Rectified Linear Unit (ReLU) activation layers,
three 3D max-pooling layers, three fully connected layers, and a
final 2-way softmax activation layer to classify the candidates from
nodules to none-nodules. Moreover, dropout layers are added after
max-pooling layers and fully-connected layers to avoid overfitting.
We initialize the parameters of the proposed 3D DCNN by the same
strategy using in the literature \cite{He15}. The detailed
architecture of the proposed 3D DCNN is illustrated in Fig.
\ref{fig:3dcnn}.

As for inputs of the proposed 3D DCNN, we firstly normalize each CT
scan with a mean of -600 and a standard deviation of -300. After
that, for each candidate, we use the center of candidate as the
centroid and then crop a $40\times40 \times24$ patch. The strategy
for data augmentation is given as follows.
\begin{itemize}
\item \textbf{Crop} For each $40\times40\times24$ patch, we crop smaller patches in the size of $36\times36\times20$ from
it, thus augmenting 125 times for each candidate.
\item \textbf{Flip} For each cropped $36\times36\times20$ patch, we flip
it from three orthogonal dimensions (coronal, sagittal and axial
position), thus finally augmenting $8\times125=1000$ times for each
candidate.
\item \textbf{Duplicate} In training process, whether a candidate is positive
or negative is decided by whether the geometric center of the
candidate locates in a nodule or not. To balance the number of
positive and negative patches in the training set, we duplicate
positive patches by $8$ times.
\end{itemize}

Note that 3D context of candidates plays an important role in
recognizing nodules due to the inherently 3D structure of CT images.
Our 3D convolutional filters, which integrate 3D local units of
previous feature maps, can `see' 3D context of candidates, whereas
traditional 2D convolutional filters only capture 2D local features
of candidates. Hence, the proposed 3D DCNN outperforms traditional
2D DCNNs.
\section{Experimental Results and Discussions}
In this section, we evaluate the performance of our CAD system on
the LUNA16 Challenge \cite{luna}. Its dataset was collected from the
largest publicly available reference database for pulmonary nodules:
the LIDC-IDRI \cite{lidc}, which contains a total of 1018 CT scans.
For the sake of pulmonary nodules detection, CT scans with slice
thickness greater than 3 mm, inconsistent slice spacing or missing
slices were excluded, leading to the final list of 888 scans. The
goal of this challenge is to automatically detect nodules in these
volumetric CT images.

In the LUNA16 challenge, performance of CAD systems are evaluated
using the Free-Response Receiver Operating Characteristic (FROC)
analysis \cite{luna}. The sensitivity is defined as the fraction of
detected true positives divided by the number of nodules. In the
FROC curve, sensitivity is plotted as a function of the average
number of false positives per scan (FPs/scan). The average
FROC-score is defined as the average of the sensitivity at seven
false positive rates: 1/8, 1/4, 1/2, 1, 2, 4, and 8 FPs per scan.
\subsection{Candidate Detection Results}
\begin{table}[htbp]
\centering \caption{The comparison of CAD systems in the task of
candidate detection.}
\begin{tabular}{p{5cm}|p{2cm}|p{3cm}}
\hline
System&~~Sensitivity &~~Candidates/scan\\
\hline
ISICAD      &~~0.856       &~~335.9 \\
SubsolidCAD &~~0.361       &~~290.6 \\
LargeCAD    &~~0.318       &~~47.6 \\
M5L         &~~0.768       &~~22.2 \\
ETROCAD     &~~0.929       &~~333.0\\
\hline
Baseline(w/o deconv) &~~0.817 &~~22.7\\
Baseline(4 anchors) &~~0.895 &~~25.8\\
\textbf{Ours}         &~~\textbf{0.946}       &~~\textbf{15.0}\\
\hline
\end{tabular}
\label{tab:tab_candi} \vspace{-0.2in}
\end{table}
The candidate detection results of our CAD system, together with
other candidate detection methods submitted to LUNA16 \cite{luna},
are shown in Table~\ref{tab:tab_candi}. From this table, we can
observe that our CAD system has achieved the highest sensitivity
(94.6 $\%$) with the fewest candidates per scan (15.0) among these
CAD systems, which verifies the superiority of the improved Faster
R-CNN in the task of candidate detection. We also make comparison to
two baseline methods of the improved Faster R-CNN (See Table
\ref{tab:tab_candi}). `Baseline(w/o deconv)' is a baseline method
where the deconvolutional layer is omitted in feature extraction and `Baseline(4 anchors)' is a baseline method where only four anchors
(i.e. $4\times4$, $10\times10$, $16\times16$, and $32\times32$) are
used in improved Faster R-CNN. According to Table
\ref{tab:tab_candi}, the comparison between `Ours' vs. `Baseline(w/o
deconv)' verifies the effectiveness of the deconvolutional layer in
the improved Faster R-CNN, while the comparison between `Ours' vs.
`Baseline(4 anchors)' indicates that the proposed 6 anchors are more
suitable for candidate detection.
\subsection{False Positive Reduction Results}
\begin{figure}[htbp]
\vspace{0.02in}
\begin{center}
\includegraphics[width=\columnwidth]{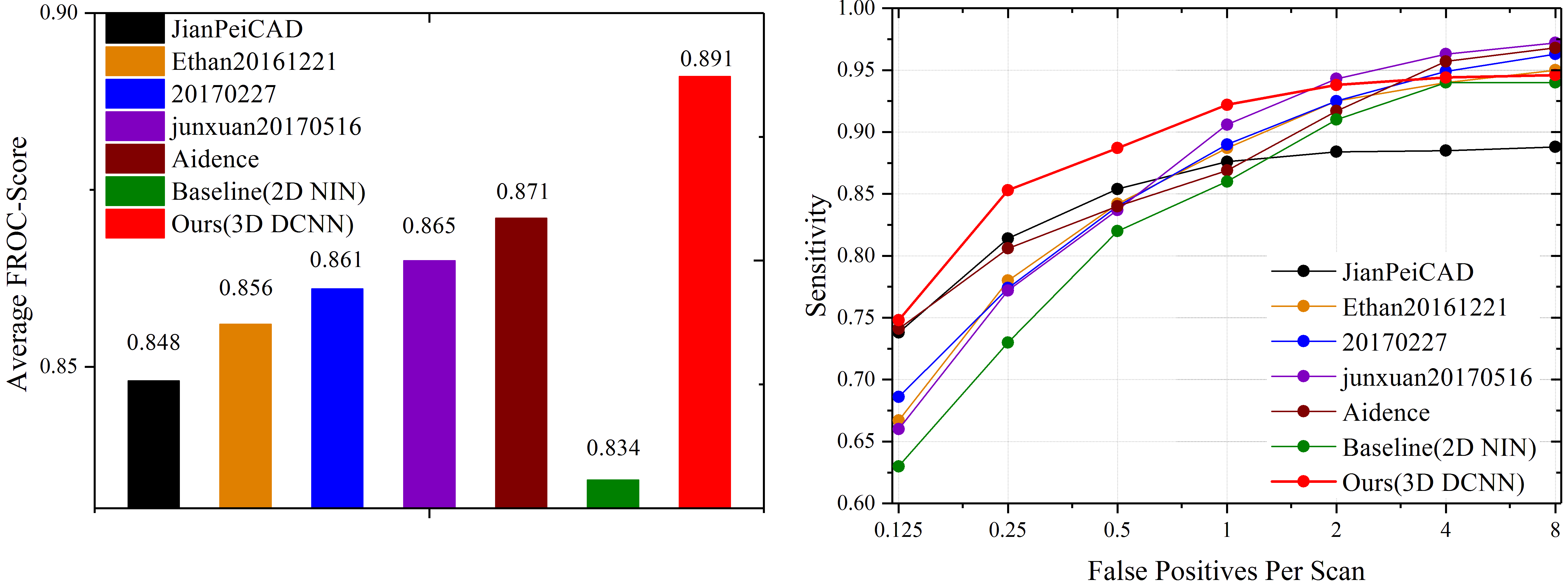}
\end{center}
\vspace{-0.2in} \caption{Comparison of performance among our CAD
system and other submitted approaches on the LUNA16 Challenge. (a)
Average FROC-scores (b) FROC curves} \label{fig:froc}
\end{figure}

To evaluate the performance of our 3D DCNN in the task of false
positive reduction, we conduct a baseline method using the NIN
\cite{nin}, a state-of-the-art 2D DCNN model for general image
recognition. To fit the NIN into the task of false positive
reduction, we modify its input size from $32 \times 32 \times 3$
into $36\times 36 \times 7$ and the number of final softmax outputs
is changed into 2. For fair comparison, we use the same candidates
and data augmentation strategy with the proposed 3D DCNN. The
comparison of the two DCNNs in the task of false positive reduction
are provided in Fig. \ref{fig:froc}. Experimental results
demonstrate that our 3D DCNN significantly outperforms 2D NIN, which
verifies the superiority of the proposed 3D DCNN to 2D DCNN in false
positive reduction.
We further present the comparison among top results on the
leaderboard of LUNA16 Challenge\footnote{Until the submission of
this paper, https://luna16.grand-challenge.org/results/}, which is
shown in Fig. \ref{fig:froc}. From this figure, we can observe that
our model has attained the best performance among the CAD systems
submitted in the task of nodule detection. Although `Ethan20161221'
and `20170227' yield comparable performance when the number of FPs/scan
is more than 2, however, they perform a significant drop with less
than 2 FPs/scan, which limits their practicability in nodule
detection. Moreover, Aidence trained its model using the labeled
data on the NLST dataset \cite{NLST}, therefore, its result is
actually incomparable to ours. Noted that, since most CAD systems
used in clinical diagnosis have their internal threshold set to
operate somewhere between 1 to 4 false positives per scan on
average, our system satisfies clinical usage perfectly.
\section{Conclusion}
In this study, we propose a novel pulmonary nodule detection CAD
system based on deep convolution networks, where a deconvolutional
improved Faster R-CNN is developed to detect nodule candidates from
axial slices and a 3D DCNN is then exploited for false positive
reduction. Experimental results on the LUNA16 Nodule Detection
Challenge demonstrate that the proposed CAD system ranks the 1st
place of Nodule Detection Track (NDET) with an average FROC-score of
0.891. We believe that our CAD system would be a very powerful tool
for clinal diagnosis of pulmonary cancer.

\bibliographystyle{plain}
\bibliography{c.bib}

\end{document}